\documentclass[journal]{IEEEtran}

\ifCLASSINFOpdf
\else
   \usepackage[dvips]{graphicx}
\fi
\usepackage{url}
\usepackage{amsmath}

\usepackage{graphicx}
\usepackage{array}
\usepackage{float}

\begin{document}

\title{Deepfake Face Traceability with Disentangling Reversing Network}

\author{Jiaxin Ai, 
Zhongyuan Wang, \IEEEmembership{Member, IEEE},
Baojin Huang, 
and Zhen Han}

\maketitle

\begin{abstract}
Deepfake face not only violates the privacy of personal identity, but also confuses the public and causes huge social harm. The current deepfake detection only stays at the level of distinguishing true and false, and cannot trace the original genuine face corresponding to the fake face, that is, it does not have the ability to trace the source of evidence. The deepfake countermeasure technology for judicial forensics urgently calls for deepfake traceability. This paper pioneers an interesting question about face deepfake, active forensics that “know it and how it happened”. Given that deepfake faces do not completely discard the features of original faces, especially facial expressions and poses, we argue that original faces can be approximately speculated from their deepfake counterparts. Correspondingly, we design a disentangling reversing network that decouples latent space features of deepfake faces under the supervision of fake-original face pair samples to infer original faces in reverse.
\end{abstract}


\IEEEpeerreviewmaketitle

\section{Introduction}
Deepfake is a technology to create or synthesize fake content (such as images, videos, and voices) based on intelligent methods such as deep learning [1]. In recent years, with the development of deep learning technology, deepfake has been developing at an unprecedented speed. At present, deepfake technology can not only generate face-swapping images and imitate the expressions of real people, but also create characters that do not exist in reality and are difficult to distinguish, subverting the traditional concept of “seeing is believing”. 

Once deepfake technology is abused, it will bring great harm to individuals, society and the country. Criminals employ deepfake technology to create fake pornographic or inappropriate speech videos to slander, frame and blackmail others. Deep forgery technology creates false information that resembles the real one, further endangering national security and public safety. For example, in the recent Russian-Ukrainian conflict, a large number of photos, videos and information have been posted on various social media, and the false and confusing information of deepfake is endless. False information that confuses the public not only intensifies social conflicts, incites violence and terrorist actions, but also becomes a psychological weapon on the battlefield. 

Although the existing deepfake detection has undergone a lot of research, it can only detect whether an image or a video is forged, but cannot trace the source of the forged information. In the case of deepfake faces, it cannot unveil the original genuine face before the deepfake. Unfortunately, the current deepfake identification method has the limitation of “knowing the truth, but not knowing the reason”, which limits its forensic power. Therefore, it is necessary to study the identification technology with traceability, and further expand the authenticity identification to the attribution of the deepfake scheme. However, to the best of our knowledge, there is currently little research in this area. 

The existing deepfake face detection is mainly engaged in authenticity discrimination, but there is little research on attribution of forgery patterns. These authentication algorithms can only verify whether the face image is forged, but cannot trace the true source of the forged face. In other words, it is impossible to determine which genuine person’s face is replaced. Due to the lack of interpretability and a chain of evidence that confirms each other in the identification conclusion, the value of judicial evidence is weakened. This paper attempts to further reverse the original face on the basis of detection, the so-called deepfake inversion.

Deepfake procedure feeds the original face into a GAN forgery model towards a specific target person, and outputs a fake face. In this paper, the three kinds of face images involved in the forgery process are called original face, target face and fake face respectively. The deepfake face is the information mixture of the original face and the target face. Although the deepfake face looks highly similar to the target face, it has more or less trace information of the original face. We can construct an inverse mapping network from fake faces to original faces, and then collect a large number of real-fake face sample pairs to train the constructed reverse mapping network, which is used to speculate the original face corresponding to a given fake face. 

The key to inverse mapping network is to disentangle the original and target facial features from the fake face blending the two. Facial feature exchange or blending of fake faces is done in the latent space of the GAN rather than the pixel domain. Face representation disentanglement in the latent space has been widely exploited [12, 13], which converts a given face back to the latent space of a pre-trained GAN model, then decouples and edits the latent features, and then inversely maps it to the image space to generate faces with changed expressions, poses, and even identity attributes. This is the so-called GAN inversion based face editing. In the light of GAN inversion, we intend to design a fake face reverse mapping model with decoupling learning, composed of a multi-branch decoupling learning structure, and formulates loss functions such as content loss and contrast loss. The face is reversely deconstructed into the original face, so as to realize the traceability and forensics of the source of the fake face.

\section{Related Work}
\label{sec:guidelines}
A lot of efforts have been put into the deepfake detection. To counter this deepfake threat, He {\em et al.} [2] constructed the ForgeryNet dataset, an extremely large face forgery dataset with unified annotations in image- and video-level data across four tasks. 

Li {\em et al.} [3] proposed to do “X-Ray” for face-swapped images, to detect whether the image is a composite picture, and to point out the boundary of the composite, which has two characteristics of recognition and interpretation. To achieve better generalization in the detection of unknown forgery methods, Yu {\em et al.} [4] proposed a commonality learning strategy for face video forgery detection to improve the generalization, which learns the common forgery features from different forgery databases, including FaceForensics, DFDC and CelebDF. 

Existing detection approaches contribute to exploring the specific artifacts in deepfake videos, but with modern deepfake technology, attackers can produce the talking video of a user without leaving any visually noticeable fake traces. Yang {\em et al.} [5] proposed a new visual speaker authentication (or face-based and lip-based authentication) scheme to defend against sophisticated deepfake attacks, which is able to achieve an accurate authentication result against human imposters. Given the empirical results that the identities behind voices and faces are often mismatched in deepfake videos, Cheng {\em et al.} [6] proposed to perform the deepfake detection from an unexplored voice-face matching view, where a voice-face matching detection model is devised to measure the matching degree of these two on a generic audio-visual dataset. However, the existing defense against deepfake is mainly focused on the identification of authenticity rather than attribution. 

Regarding the reverse traceability of deepfake, the only attempts at present are only from the perspective of digital watermarking or blockchain contracts. Hasan {\em et al.} [7] constructed a deepfake video detection method based on blockchain and smart contracts. They associate each video with a smart contract that is linked to its parent video, where each parent video has a child video linked to its hierarchy. With this link, users can reliably trace them back to the original video, even if the video has been copied multiple times. But the premise is that the video is only considered true if the source is traceable. In fact, there are many videos from unknown sources that are not all fake videos. There are also some studies that explore the generation model of fake faces from the perspective of GAN inversion [8]. For example, Yu {\em et al.} [9] proposed a classification method of GAN fingerprints, which can not only distinguish between real images and generated images, but also effectively identify the models from which different generated images are derived. Asnani {\em et al.} [10] developed a deepfake reverse engineering method that not only identifies a fake face, but also infers the machine learning model that created it, including the network architecture and training loss function. However, these two studies only focus on the inversion of GAN models that generate fake or synthetic images rather than fake images, and alternatively, they cannot be used to reverse infer original faces corresponding to fake faces. In a further inspection, such methods are based on the fact that images generated by different GANs have unique features in the intermediate classification layer, which can be used as the discriminative fingerprint of the GAN generator. Therefore, if the GAN fingerprint is deliberately destroyed during the deepfake process, it is difficult for these methods to work anymore. Neves {\em et al.} [11] presented a strategy to remove GAN fingerprints from synthetic fake images in order to spoof facial manipulation detection systems while keeping the visual quality of the resulting images.

\section{Proposed Mapping Model}
\subsection{Architecture}
Given a fake face \(I_{fake}\) , our goal is to generate a traced face \(I_{tra}\)  with the same identity from the original face \(I_{ori}\). We divide the objective into two parts, disentangling and reversing, and tackle them separately. The former disentangles intrinsic identity features from identity-independent attributes, such as pose, expression and illumination. The latter infers original facial identities from fake facial identities. Once achieved, we can extract features from \(I_{fake}\)  and disentangle them into identity and attributes. Then we infer the original facial identity from the fake facial identity, concatenating it with fake facial attributes, and feed the combined representation into a decoder to generate \(I_{tra}\). Following this idea, our proposed deepfake face traceability network consists of identity disentangling and reversing modules, as shown in Fig. 1 (a).

\textbf{Identity disentangling module.} It consists of an identity encoder \(E_{id\_ori}\), an attribute encoder \(E_{attr}\) and a shared decoder \(D\). Given an original face \(I_{ori}\)  as input, \(E_{id\_ori}\) and \(E_{attr}\) are used to extract identities and attributes from \(I_{ori}\) respectively. To disentangle identities from all other attributes, we supervise \(E_{id\_ori}\)  with the output of pretrained face recognition model of high accuracy, while minimizing the redundancy between the outputs of the two encoders. To enforce disentanglement cycle consistency, we concatenate the extracted identity and attribute features in channel dimension and feed them into \(D\) to reconstruct the original face. 

\textbf{Face reversing module.} The task of generating \(I_{tra}\) with the same identity as in \(I_{ori}\) and attributes portrayed in \(I_{fake}\) consists of two parts: \textbf{1)} extracting original facial identities and fake facial attributes from \(I_{fake}\) and \textbf{2)} assembling them to create a new representation and generating an image accordingly. For the former, we use the original facial identity encoder \(E_{id\_ori}\) and the attribute encoder \(E_{attr}\) to extract identities and attributes from \(I_{fake}\) respectively. We supervise the extracted identities by original facial identities to learn the mapping from fake facial identities to original facial identities. Meanwhile, since \(I_{fake}\) usually shares the same attributes with \(I_{ori}\), we directly employ attributes extracted from \(I_{ori}\) to constrain the extracted attributes of \(I_{fake}\). For the latter, we concatenate the extracted identities and attributes and feed them into the shared decoder \(D\), which then generates the resulting face. At the end, we use the original facial identity encoder to extract identity features from \(I_{tra}\), which is used for calculating loss to enforce cycle consistency and identity preservation.

\begin{figure*}
\centerline{\includegraphics[width=\linewidth]{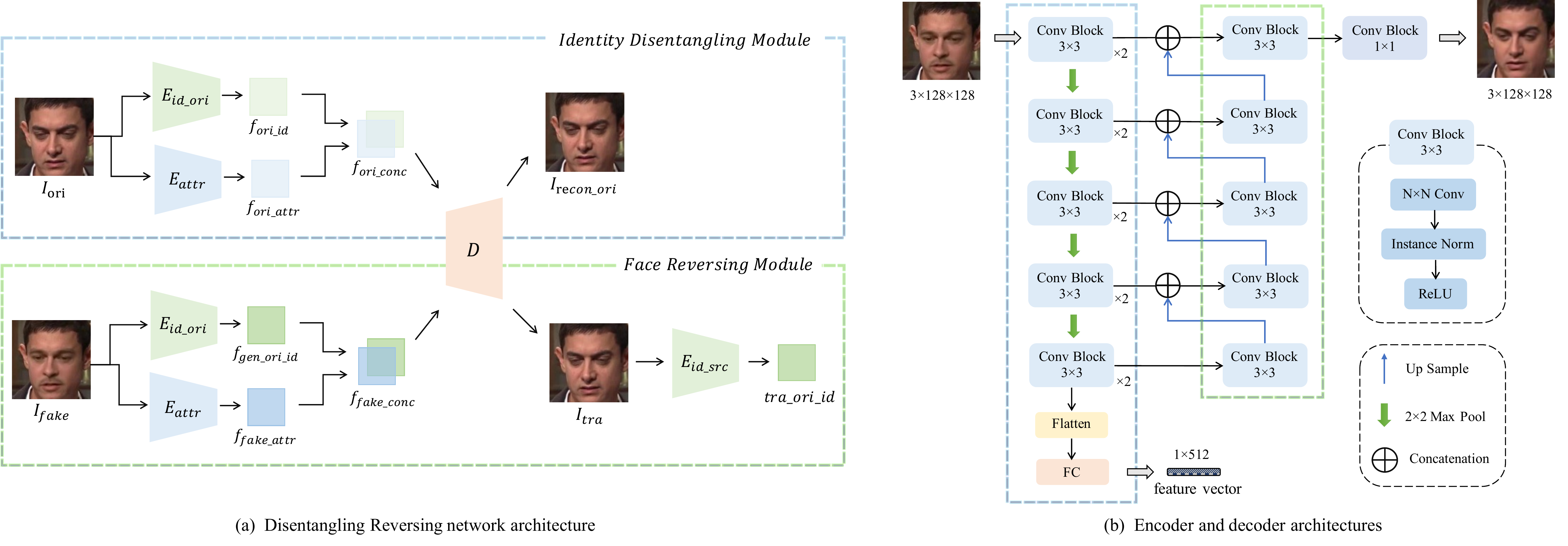}}
\vspace{-0.08cm}
\caption{(a) shows the disentangling reversing network architecture. The solid black lines represent data flow. In identity disentangling module above, the original facial identity and attribute features \(f_{ori\_id}\), \(f_{ori\_attr}\) are first extracted by original facial identity encoder \(E_{id\_ori}\) and attribute encoder \(E_{attr}\), respectively. Then we combine \(f_{ori\_id}\) and \(f_{ori\_attr}\) into \(f_{ori\_conc}\) and feed it into a shared decoder \(D\) to reconstruct original faces. In face reversing module, the generated original identity feature \(f_{gen\_ori\_id}\) is first extracted from the fake face. Then we assemble it with fake attribute feature \(f_{fake\_attr}\) extracted by \(E_{attr}\) and feed the concatenated feature \(f_{fake\_conc}\) into to generate traced face with the same identity in original face. The traced identity \(f_{tra\_id}\) is also extracted for loss calculation to enforce cycle consistency. (b) shows the architectures of our encoders \(E_{id\_ori}\), \(E_{attr}\) and decoder \(D\) based on UNet. The encoder takes a 3$\times$128$\times$128 input to generate a 1$\times$512$\times$16$\times$16 feature map, which is then fed to the decoder to reconstruct image after 4 up-sampling steps. We adopt the skip-connection in UNet to concatenate feature maps of the same scale. Particularly, we add a flatten layer and a successive FC layer to generate feature vectors from feature maps for identity and redundancy supervision. Besides, the decoder takes 4 up-sampling steps while we actually employ 5 up-sampling steps in training because the decoder receives doubled feature maps due to the concatenation of identity and attribute.}
\label{fig:1}
\end{figure*}

The structures of \(E_{id\_ori}\), \(E_{attr}\) and \(D\) are shown in Fig. 1 (b), based on UNet [14]. The reason for adopting a similar UNet structure lies in the following two points. First, the encoder and decoder in UNet have relatively shallow layers. Since our network consists of 3 models, it is conductive to have a shallow structure to control the parameters and prevent the models from being too large to apply. Besides, UNet adopts a skip-connection strategy to concatenate feature maps obtained by up-sampling path with those by down-sampling path in the same scale, which ensures the finally recovered feature maps to integrate more low-level features and fuse different scales of features, and thus to reconstruct more realistic faces. 

Different from the completely symmetrical U-shaped structure of UNet, we have made some modifications. First, we add a flatten layer and a full connection layer after the four down-sampling steps of encoder, so as to convert the face identity and attribute feature maps into feature vectors. Second, our identity and attribute encoders are both structured with four layers, while the shared decoder is structured with five layers. This is because the decoder receives doubled feature maps after concatenation of identities and attributes. Moreover, while concatenating feature maps extracted from identity and attribute encoders as mentioned above, we actually concatenate the feature maps obtained in each down-sampling step and feed to the corresponding double-scale up-sampling step, so as to adopt the skip-connection strategy. 

\subsection{Optimization}
The identity disentangling module takes original face image \(I_{ori}\) as input, which aims at disentangling the intrinsic identity from all other attributes and assemble the decoupled features to reconstruct the original face \(I_{recon\_ori}\).

Firstly, to enforce the identity encoder’s ability of extracting benign original facial identity features from \(I_{ori}\), we employ \(l_1\) distance between the output of \(E_{id\_ori}\) and extracted original facial identity features using a pretrained ResNet-50 face recognition network \(E_{id\_pre}\). 
\begin{equation}
  \setlength{\abovedisplayskip}{5pt}
  \setlength{\belowdisplayskip}{5pt}
  L_{id} = \left\| E_{id\_ori}(I_{ori})-E_{id\_pre}(I_{ori}) \right\|_1
  \label{eq:loss_id}
\end{equation}

Secondly, we adopt a cosine similarity to reduce the redundancy between identity features and attribute features as much as possible. 
\begin{equation}
\setlength{\abovedisplayskip}{5.5pt}
\setlength{\belowdisplayskip}{6pt}
\begin{aligned}
    L_{redun}
    &= \frac{E_{id\_ori}(I_{ori})^TE_{attr}(I_{ori})}{\left\| E_{id\_ori}(I_{ori})\right\|^2_2 \cdot \left\| E_{attr}(I_{ori}) \right\|^2_2} \\
    &+ \frac{E_{id\_ori}(I_{fake})^TE_{attr}(I_{fake})}{\left\| E_{id\_ori}(I_{fake})\right\|^2_2 \cdot \left\| E_{attr}(I_{fake}) \right\|^2_2}
    \label{eq:loss_redun}
\end{aligned}
\end{equation}

In addition, a reconstruction loss is exercised to encourage pixel-level fidelity between \(I_{recon\_ori}\) and \(I_{ori}\).
\begin{equation}
\setlength{\abovedisplayskip}{5pt}
\setlength{\belowdisplayskip}{5pt}
  L_{recon} = \left\| I_{ori}-I_{recon\_ori} \right\|^2_2
  \label{eq:loss_recon}
\end{equation}

The reversing module takes fake face image \(I_{fake}\) as input, which intends to learn the mapping from the fake facial identities to original facial identities, and generate the face \(I_{tra}\) as similar to \(I_{fake}\) as possible.

Firstly, we adopt \(l_1\)  distance between identity features extracted from \(I_{fake}\)  and original facial identities to learn the mapping relationship. 
\begin{equation}
\setlength{\abovedisplayskip}{5pt}
\setlength{\belowdisplayskip}{5pt}
  L_{map} = \left\| E_{id\_ori}(I_{ori})-E_{id\_ori}(I_{fake}) \right\|_1
  \label{eq:loss_map}
\end{equation}

Besides, we minimize \(l_2\)  distance between \(I_{tra}\) and \(I_{ori}\) to encourage  \(I_{tra}\) similar to \(I_{ori}\).
\begin{equation}
\setlength{\abovedisplayskip}{5pt}
\setlength{\belowdisplayskip}{5pt}
  L_{gen} = \left\| I_{ori}-I_{tra} \right\|^2_2
  \label{eq:loss_gen}
\end{equation}

Finally, \(l_1\) distance between identities of \(I_{tra}\) and \(I_{ori}\) is used to enforce cycle consistency and identity preservation. 
\begin{equation}
\setlength{\abovedisplayskip}{5pt}
\setlength{\belowdisplayskip}{5pt}
  L_{cycle} = \left\| E_{id\_ori}(I_{ori})-E_{id\_ori}(I_{tra}) \right\|_1
  \label{eq:loss_map}
\end{equation}

The overall loss is a weighted combination of the described losses, formulated as:
\begin{equation}
\setlength{\abovedisplayskip}{5.5pt}
\setlength{\belowdisplayskip}{5.5pt}
\begin{aligned}
  L 
  &=\lambda_1L_{id}+\lambda_2L_{redun}+\lambda_3L_{recon} \\
  &+\lambda_4L_{map}+\lambda_5L_{gen}+\lambda_6L_{cycle}
  \label{eq:loss}
\end{aligned}
\end{equation}

\section{Experiments}
We conducted experiments on Celeb-DF-v2 [15] and FaceForensics++ [16] datasets to examine the quality of speculated faces. Since this is the first work on deepfake face reversion, there are no existing comparison baselines. 

\subsection{Experimental Settings}
\textbf{Datasets.} Celeb-DF-v2 has 590 original videos covering 62 celebrities and 5639 manipulated videos generated by swapping faces for each pair of the 62 subjects. FaceForensics++ contains 28 subjects, with 1000 original videos and 3000 manipulated videos. 

\textbf{Training details.} We perform temporal down-sampling on the two datasets by selecting one frame every second in each video, and obtain a total number of 74,118 pairs of fake and original faces in Celeb-DF-v2 and 29,388 pairs in FaceForensics++. Among them, we randomly select 67,513 pairs and 26,506 pairs as the training sets respectively for Celeb-DF-v2 and FaceForensics++. The rest samples are used for testing. 

The disentangling reversing network was trained on 128$\times$128 pixels to give outputs of the same size. It minimized the loss function given by Eq. (7), where hyper-parameters are set to 1, except that the weight \(\lambda_6\) is set to 5. On both datasets, the training was performed using Adam for 200 iterations with a batch size of 32. The momentums were set to 0.5 and 0.999 and the learning rate was initialized to 0.0003. LeakyReLU and Kaiming method [17] were employed for nonlinearity and weight initialization, respectively. The training reached convergence after 140 iterations on Celeb-DF-v2 and 80 iterations on FaceForensics++.

\begin{figure}[H]
\centerline{\includegraphics[width=\columnwidth]{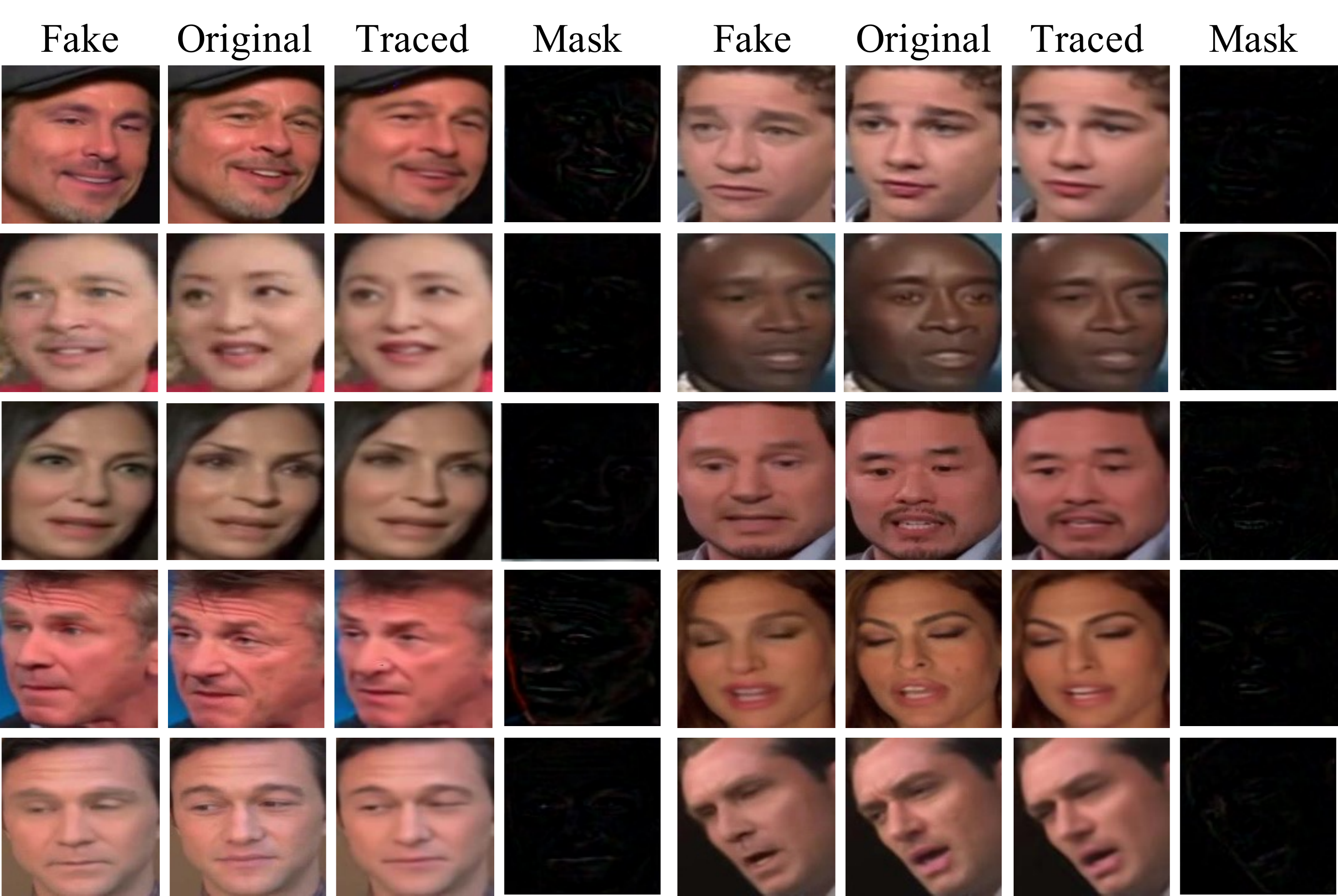}}
\vspace{-0.08cm}
\caption{Visual results on Celeb-DF-v2. The speculated faces from the fake faces look highly similar to the original genuine faces.}
\label{fig:2}
\end{figure}

\subsection{Results}
The visual results are shown in Fig. 2 and 3 for Celeb-DF-v2 and FaceForensics++, respectively. The mask image shows the difference between original and traced faces. The greater the intensity of the mask, the greater the gap between the inverted face and the original face; and vice versa.

The results qualitatively demonstrate that our network successfully reverses original faces from fake faces provided that traced faces are almost indistinguishable from original faces visually. Besides, we observe that when fake and original face images have different facial expressions, the traced faces retain attributes in fake faces. This verifies our network’s ability to disentangle identities from other attributes, and generate traced faces with the same attributes in fake faces. 
Further, we quantitatively evaluate the tracing performance of our network using several metrics, such as PSNR, SSIM, as well as facial similarities between original and traced faces. Among them, we extract facial features through Arcface [18] recognizer and then calculate the cosine similarity of the features. Relatively, the facial similarity metric given by face recognizer can better indicate the similarity of faces in the sense of identity. The results on both datasets are tabulated in Table I, which quantitatively show the high fidelity of speculated faces to original faces.

\begin{figure}
\centerline{\includegraphics[width=\columnwidth]{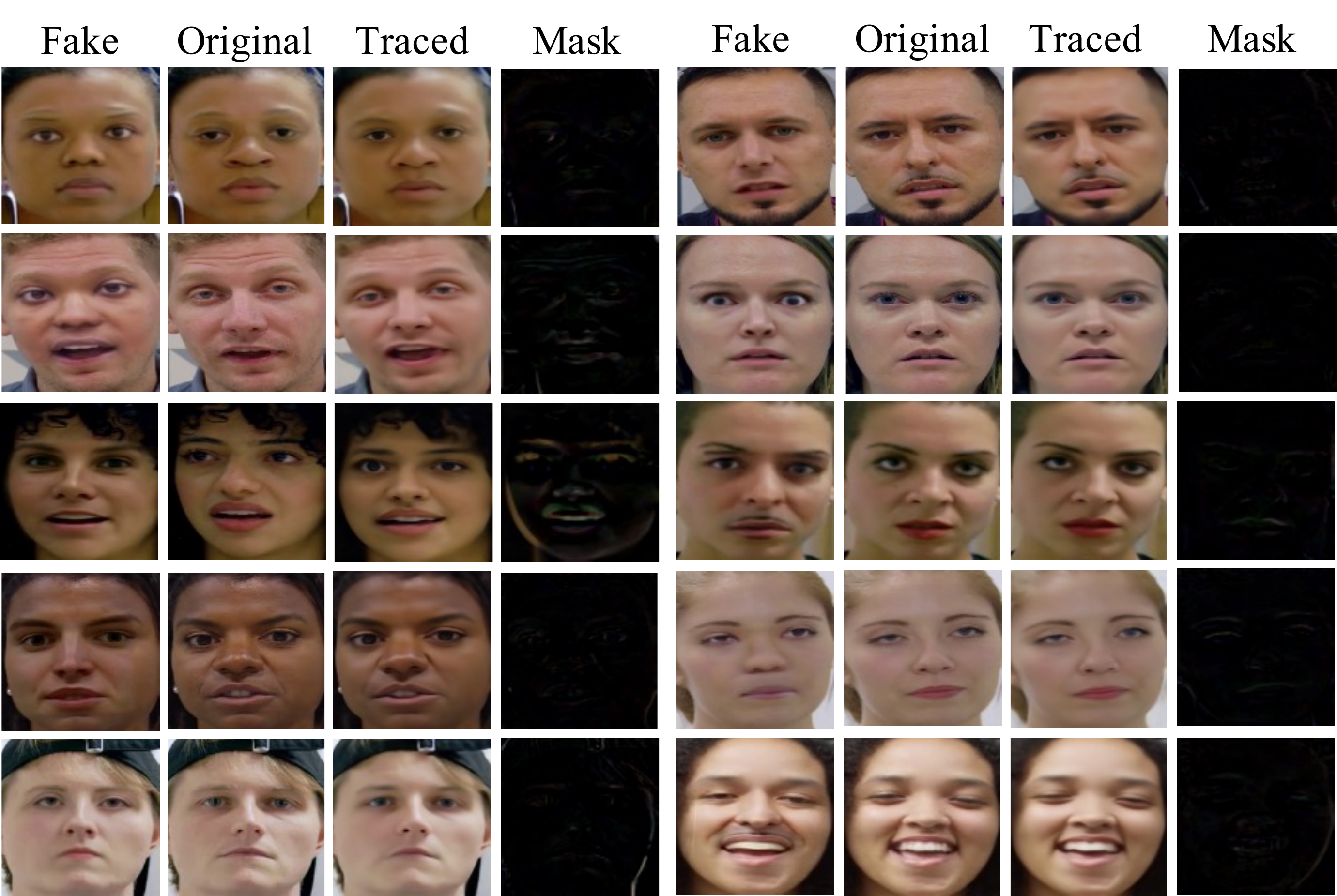}}
\vspace{-0.08cm}
\caption{Visual results on FaceForensics++. The speculated faces from the fake faces look highly similar to the original genuine faces.}
\label{fig:3}
\end{figure}

We notice that quantitative results on FaceForensics++ are generally worse than on Celeb-DF-v2. We additionally show a few failure cases on FaceForensics++ in Fig. 4. As seen, our network cannot well handle the samples with large differences in skin color or gender between fake and original faces. In addition, the quality of fake faces also has great impact on the reversion. It is difficult for our network to extract effective information from low-quality fake faces, which thus results in poor effects for speculated faces.

\begin{table}
\caption{Quantitative results on Celeb-DF-v2 and FaceForensics++}
\vspace{-0.08cm}
\label{tab:1}
\small
\setlength{\tabcolsep}{3pt}
\begin{tabular}{|p{70pt}|p{45pt}<{\centering}|p{30pt}<{\centering}|p{85pt}<{\centering}|}
\hline
Dataset & PSNR (dB) & SSIM & Facial Similarity (\%) \\
\hline
Celeb-DF-v2 & 33.16 & 0.9008 & 81.17 \\
FaceForensics++ & 30.60 & 0.7710 & 71.32 \\
\hline
\end{tabular}
\end{table}

\begin{figure}[H]
\centerline{\includegraphics[width=\columnwidth]{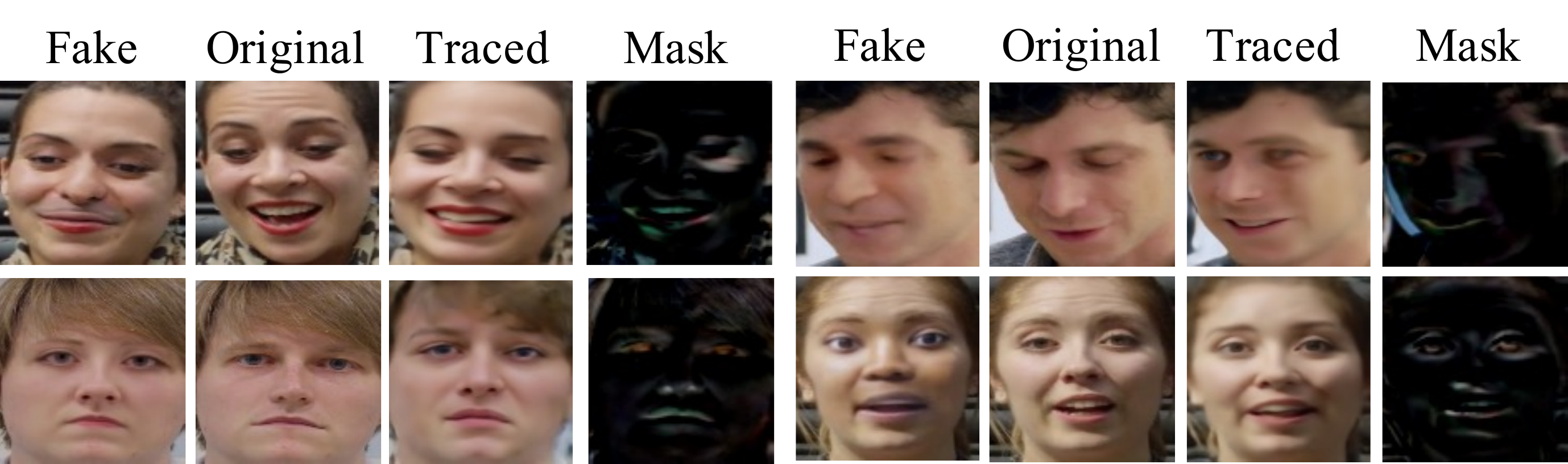}}
\vspace{-0.08cm}
\caption{A few failure cases on FaceForensics++. The mask images show relatively huge differences between original and traced images. This is mainly due to certain difficult variations, such as gender differences (left half), and poor fake quality (right half).}
\label{fig:4}
\end{figure}

\section{Conclusions}
This paper advocates a new and interesting task, deepfake traceability rather than detection. Particularly, we also propose a disentangling reversing network to accomplish this task. Experimental results show that our network can precisely reverse the original face images from the fake counterparts. Our work may spark a completely different topic in deepfake than in the past.

\section{Acknowledgements}
This work was supported by the National Key Research and Development Program of China (2021YFF0602102) and National Natural Science Foundation of China (U1903214, 62072347, 62071339).

\end{document}